\documentclass{IEEEtran}
\usepackage{cite}
\usepackage{amsmath,amssymb,amsfonts}
\usepackage{algorithmic}
\usepackage{graphicx}
\usepackage{hyperref}
\usepackage{textcomp}
\usepackage{pdflscape}
\usepackage[table]{xcolor}
\usepackage{tikz,pgfplots}
\usepackage{subcaption}
\usepackage{rotating}
\usepackage{balance}
\usetikzlibrary{pgfplots.dateplot}
\usepackage{pgfplotstable}
\def\BibTeX{{\rm B\kern-.05em{\sc i\kern-.025em b}\kern-.08em
    T\kern-.1667em\lower.7ex\hbox{E}\kern-.125emX}}
\pgfplotsset{compat=1.17}

\IEEEpubid{\begin{minipage}{\textwidth}\ \\ \\ \\ \\ [12pt]
 Preprint submitted to IET Computer Vision.
\end{minipage}}

\begin{document}

\title{A Survey of Modern Deep Learning based Object Detection Models}

\author{Syed Sahil Abbas Zaidi, Mohammad Samar Ansari, Asra Aslam,  \\ Nadia Kanwal, Mamoona Asghar,  and  Brian Lee 
\thanks{S.S.A. Zaidi, N. Kanwal, M Asghar and B. Lee are with the Athlone Institute of Technology, Ireland. M.S. Ansari is with the Aligarh Muslim University, India. A. Aslam is with the Insight Center for Data Analytics, National University of Ireland, Galway.  (Emails: sahilzaidi78@gmail.com, samar.ansari@zhect.ac.in, asra.aslam@insight-centre.org, nkanwal@ait.ie, masghar@ait.ie, blee@ait.ie) 
}}

\maketitle

\begin{abstract}
Object Detection is the task of classification and localization of objects in an image or video. It has gained prominence in recent years due to its widespread applications. This article surveys recent developments in deep learning based object detectors. Concise overview of benchmark datasets and evaluation metrics used in detection is also provided along with some of the prominent backbone architectures used in recognition tasks. It also covers contemporary lightweight classification models used on edge devices. Lastly, we compare the performances of these architectures on multiple metrics.
\end{abstract}


\section{Introduction}

Object detection is a trivial task for humans. A few months old child can start recognizing common objects, however teaching it to the computer has been an uphill task until the turn of the last decade. It entails identifying and localizing all instances of an object (like cars, humans, street signs, etc.) within the field of view. Similarly, other tasks like classification, segmentation, motion estimation, scene understanding, etc, have been the fundamental problems in computer vision. 

Early object detection models were built as an ensemble of hand-crafted feature extractors such as Viola-Jones detector \cite{viola_rapid_2001}, Histogram of Oriented Gradients (HOG) \cite{dalal_histograms_2005} etc. These models were slow, inaccurate and performed poorly on unfamiliar datasets. The re-introduction of convolutional neural network (CNNs) and deep learning for image classification changed the landscape of visual perception. Its use in the ImageNet Large Scale Visual Recognition Challenge (ILSVRC) 2012 challenge by AlexNet \cite{NIPS2012_c399862d} inspired further research of its application in computer vision. Today, object detection finds application from self-driving cars and identity detection to security and medical uses. In recent years, it has seen exponential growth with rapid development of new tools and techniques.

\begin{figure}[tb]
\centerline{\includegraphics[width=0.499\textwidth]{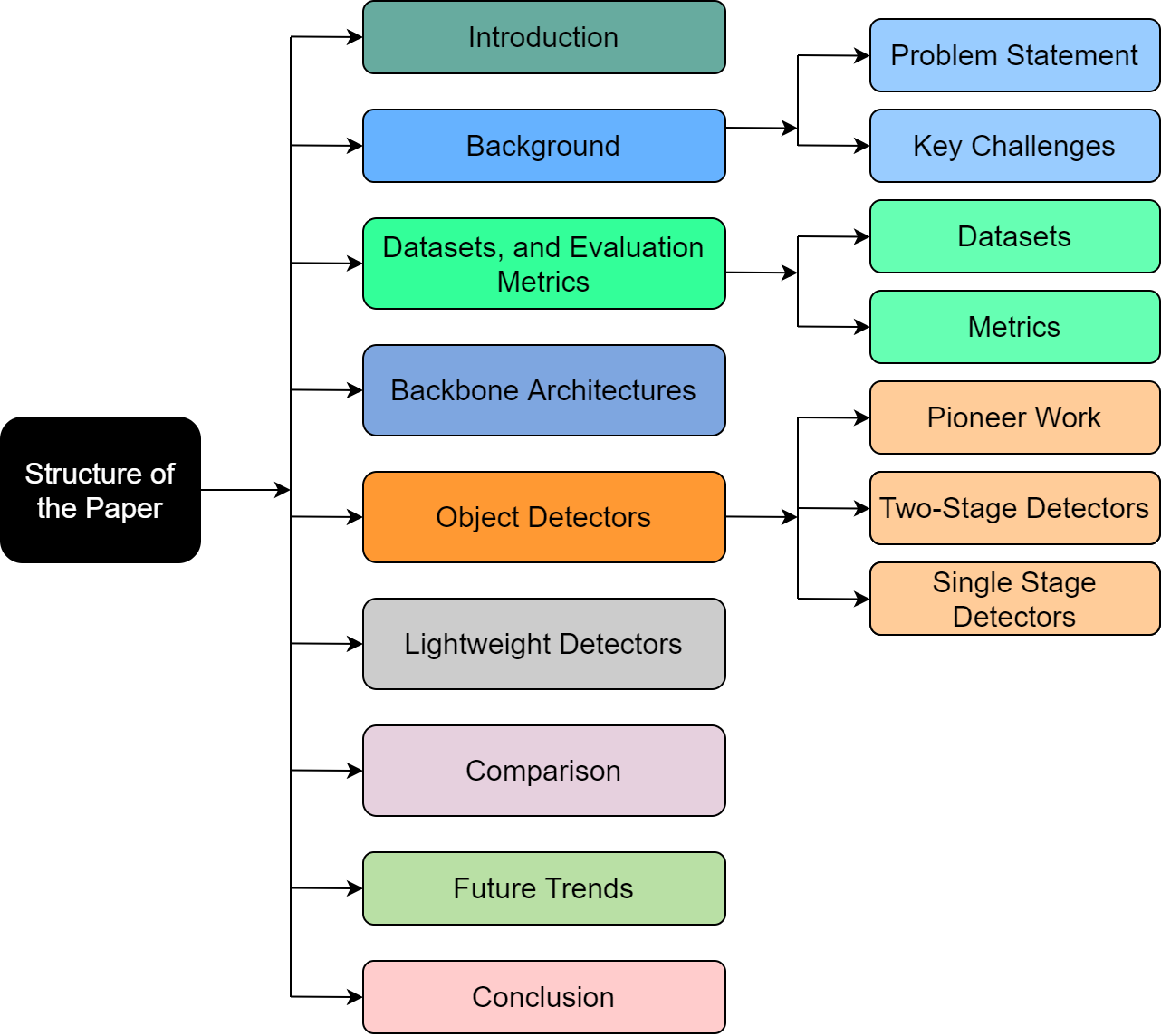}}
\caption{Structure of the paper.}
\label{fig:structure}
\end{figure}

This survey provides a comprehensive review of deep learning based object detectors and lightweight classification architectures. While existing reviews are quite thorough \cite{zou_object_2019,liu_deep_2018,chahal_survey_2018,jiao_survey_2019}, most of them lack new developments in the domain. The main contributions of this paper are as follows: 
\begin{enumerate}
\item This paper provides an in-depth analysis of major object detectors in both categories – single and two stage detectors. Furthermore, we take historic look at the evolution of these methods. 
\item We present a detailed evaluation of the landmark backbone architectures and lightweight models. We could not find any paper which provides a broad overview of both these topics. 	
\end{enumerate}
In this paper, we have systematically reviewed various object detection architectures and its associated technologies, as illustrated in figure \ref{fig:structure}. Rest of this paper is organized as follows. In section \ref{bb}, the problem of object detection and its associated challenges are discussed. Various benchmark datasets and evaluation metrics are listed in Section \ref{dm}. In Section \ref{BAr}, several milestone backbone architectures used in modern object detectors are examined. Section \ref{ObD} is divided into three major sub-section, each studying a different category of object detectors. This is followed by the analysis of a special classification of object detectors, called lightweight networks in section \ref{LtW} and a comparative analysis in Section \ref{CoR}. The future trends are mentioned in Section \ref{FuT} while the paper is concluded in Section \ref{Con}.

\begin{figure*}
     \centering
     \begin{subfigure}[b]{0.24\textwidth}
         \centering
         \includegraphics[scale = 0.76]{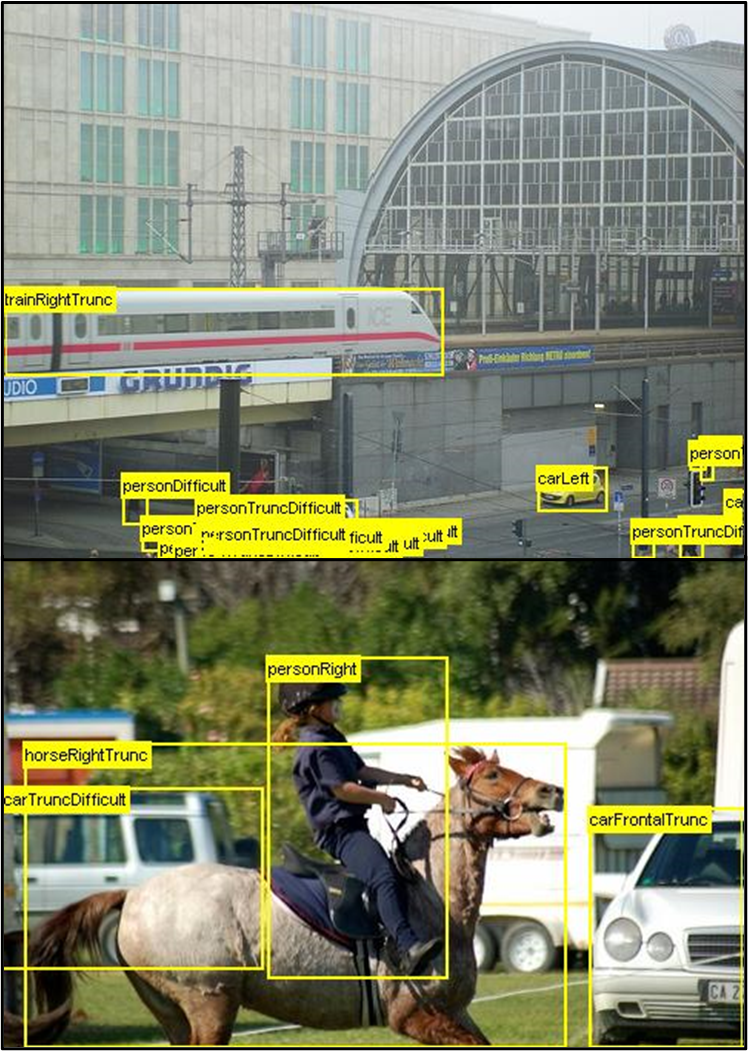}
         \caption{PASCAL VOC 12}
     \end{subfigure}
     \hfill
     \begin{subfigure}[b]{0.24\textwidth}
         \centering
         \includegraphics[scale = 0.76]{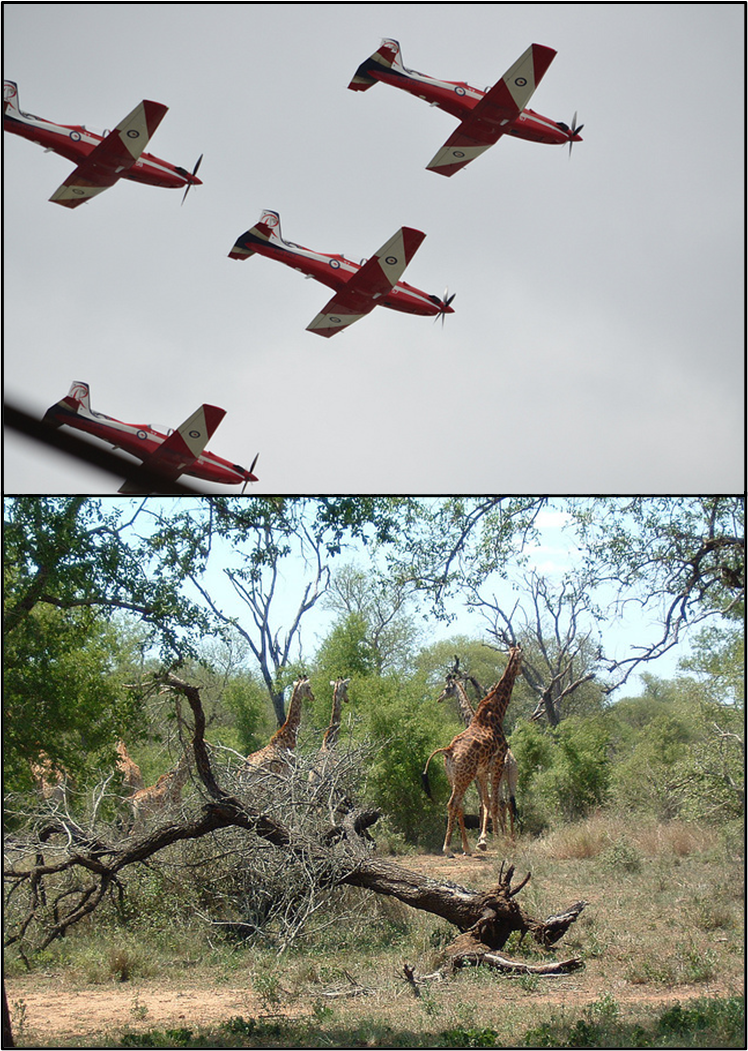}
         \caption{MS-COCO}
     \end{subfigure}
     \hfill
     \begin{subfigure}[b]{0.24\textwidth}
         \centering
         \includegraphics[scale = 0.76]{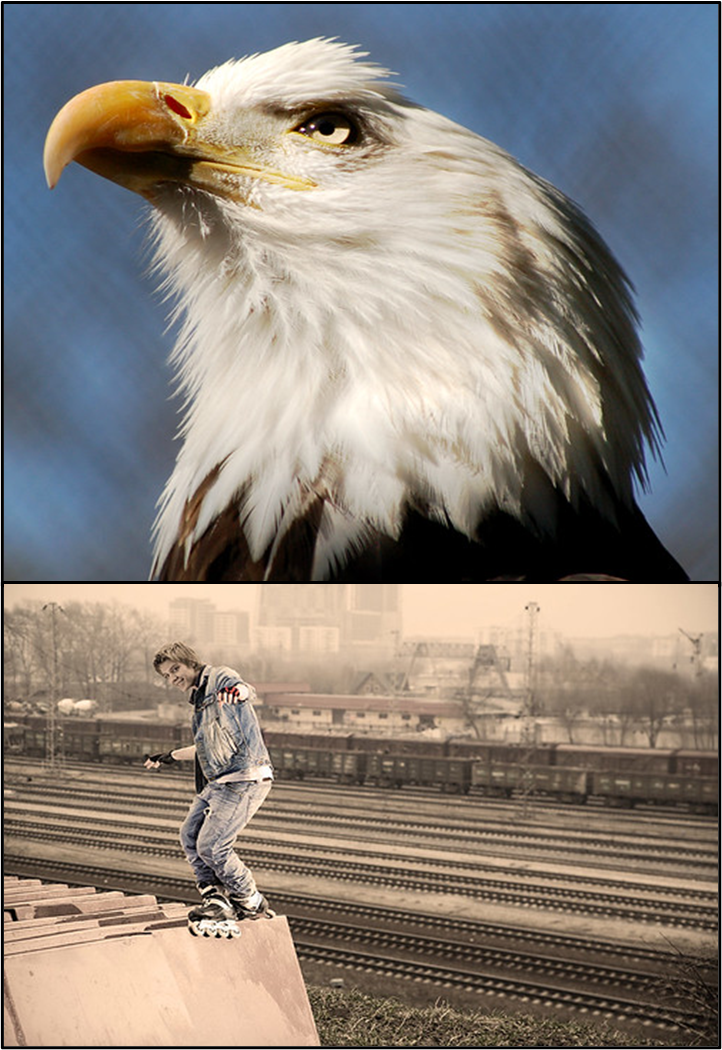}
         \caption{ILSVRC}
     \end{subfigure}
     \hfill
     \begin{subfigure}[b]{0.24\textwidth}
         \centering
         \includegraphics[scale = 0.76]{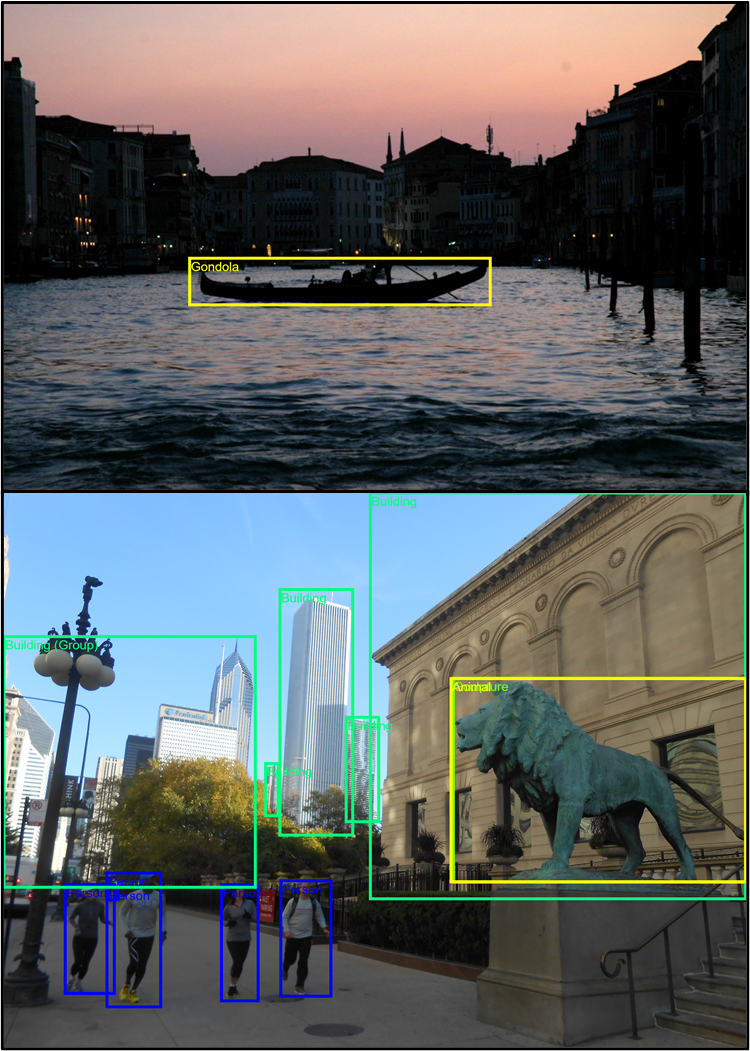}
         \caption{OpenImage}
     \end{subfigure}
        \caption{Sample images from different datasets.}
        \label{fig:sample_images}
\end{figure*}

\section{Background}
\label{bb}

\subsection{Problem Statement}
The object detection is the natural extension of object classification, which aims only at recognizing the object in the image. The goal of the object detection is to detect all instances of the predefined classes and provide its coarse localization in the image by axis-aligned boxes. The detector should be able to identify all instances of the object classes and draw bounding box around it. It is generally seen as a supervised learning problem. Modern object detection models have access to large sets of labelled images for training and are evaluated on various canonical benchmarks. 

\subsection{Key challenges in Object Detection}

Computer vision has come a long way in the past decade, however it still has some major challenges to overcome. Some of these key challenges faced by the networks in real life applications are:
\begin{itemize}
\item \textit{Intra class variation}
: Intra class variation between the instances of same object is relatively common in nature. This variation could be due to various reasons like occlusion, illumination, pose, viewpoint, etc. These unconstrained external can have dramatic effect of the object appearance \cite{liu_deep_2018}. It is expected that the objects could have non-rigid deformation or be rotated, scaled or blurry. Some objects could have inconspicuous surroundings, making the extraction difficult.

\item \textit{Number of categories}: The sheer number of object classes available to classify makes it a challenging problem to solve. It also requires more high-quality annotated data, which is hard to come by. Using fewer examples for training a detector is an open research question.
\item \textit{Efficiency}: Present day models need high computation resources to generate accurate detection results. With mobile and edge devices becoming common place, efficient object detectors are crucial for further development in the field of computer vision.
\end{itemize}

\section{Datasets and Evaluation Metrics}
\label{dm}

\subsection{Datasets}
This section presents an overview of the datasets that are available, and have been most commonly used for object detection tasks.

\subsubsection{PASCAL VOC 07/12}
\label{VOC}

The Pascal Visual Object Classes (VOC) challenge was a multiyear effort to accelerate the development in the field of visual perception. It started in 2005 with classification and detection tasks on four object classes \cite{everingham_pascal_2010}, but two versions of this challenges are mostly used as a standard benchmark. While the VOC07 challenge had 5k training images and more than 12k labelled objects \cite{pascal-voc-2007}, the VOC12 challenge increased them to 11k training images and more than 27k labelled objects \cite{everingham_pascal_2012}. Object classes was expanded to 20 categories and the tasks like segmentation and action detection were included as well. Pascal VOC introduced the mean Average Precision (\textit{mAP}) at 0.5 IoU (Intersection over Union) to evaluate the performance of the models. Figure~\ref{fig:pascalvoc_classes} depicts the distribution of the number of images w.r.t. to the different classes in the Pascal VOC dataset.

\subsubsection{ILSVRC}\label{ILS}

The ImageNet Large Scale Visual Recognition Challenge (ILSVRC) \cite{russakovsky_imagenet_2015} was an annual challenge running from 2010 to 2017 and became a benchmark for evaluating algorithm performance. The dataset size was scaled up to more than a million images consisting of 1000 object classification classes. 200 of these classes were hand-picked for object detection task, constitute of more than 500k images. Various sources including ImageNet \cite{deng_imagenet_2009} and Flikr, were used to construct detection dataset. ILSVRC also updated the evaluation metric by loosening the IoU threshold to help include smaller object detection. Figure~\ref{fig:imagenet_classes} depicts the distribution of the number of images w.r.t. to the different classes in the ImageNet dataset.

\subsubsection{MS-COCO}
\label{MSC}

\begin{figure*}[t]
\includegraphics[width=0.99\textwidth]{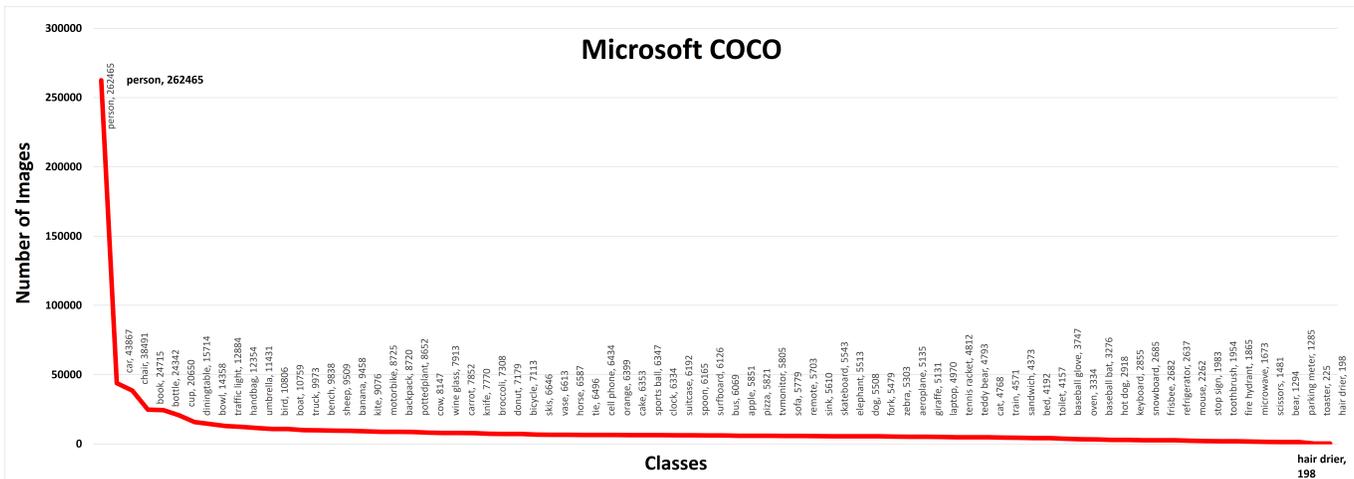}
\caption{(This image is best viewed in PDF form with magnification) Number of images for different classes annotated in the MS-COCO dataset \cite{aslam2021survey}}
\label{fig:mscoco_classes}
\end{figure*}

The Microsoft Common Objects in Context (MS-COCO) \cite{Lin_ms_coco_2014} is one of the most challenging datasets available. It has 91 common objects found in their natural context which a 4-year-old human can easily recognize. It was launched in 2015 and its popularity has only increased since then. It has more than two million instances and an average of 3.5 categories per images. Furthermore, it contains 7.7 instances per image, comfortably more than other popular datasets. MS COCO comprises of images from varied viewpoints as well. It also introduced a more stringent method to measure the performance of the detector. Unlike the Pascal VOC and ILSVCR, it calculates the IoU from 0.5 to 0.95 in steps of 0.5, then using a combination of these 10 values as final metric, called Average Precision (AP). Apart from this, it also utilizes AP for small, medium and large objects separately to compare performance at different scales. Figure~\ref{fig:mscoco_classes} depicts the distribution of the number of images w.r.t. to the different classes in the MS-COCO dataset.

\begin{table*}[]
\centering
\caption{Comparison of various object detection datasets.}
\begin{tabular}{|c|c|c|c|c|c|c|c|c|}
    \hline
    \multicolumn{1}{|c|}{Dataset}
    & \multicolumn{1}{|c|}{Classes}
    & \multicolumn{3}{c|}{Train}
    & \multicolumn{3}{c|}{Validation} 
    & \multicolumn{1}{c|}{Test} \\ 
    \cline{3-8}
    \multicolumn{1}{|c|}{}
    & \multicolumn{1}{|c|}{}
    & Images & Objects & Objects/Image
    & Images & Objects & Objects/Image
    & \multicolumn{1}{|c|}{}\\ \hline
    PASCAL VOC 12 & 20 & 5,717 & 13,609 & 2.38 & 5,823 & 13,841 & 2.37 & 10,991 \\
    MS-COCO & 80 & 118,287 & 860,001 & 7.27 & 5,000 & 36,781 & 7.35 & 40,670 \\
    ILSVRC & 200 & 456,567 & 478,807 & 1.05 & 20,121 & 55,501 & 2.76 & 40,152 \\
    OpenImage & 600 & 1,743,042 & 14,610,229  & 8.38 & 41,620 & 204,621 & 4.92 & 125,436 \\
    \hline
    \end{tabular}
    \label{table:4}
\end{table*}

\subsubsection{Open Image}
\label{OI}

Google’s Open Images \cite{kuznetsova_open_2020} dataset is composed of 9.2 million images, annotated with image-level labels, object bounding boxes, and segmentation masks, among others. It was launched in 2017 and has received six updates. For object detection, Open Images has 16 million bounding boxes for 600 categories on 1.9 million images, which makes it the largest dataset of object localization. Its creators took extra care to choose interesting, complex and diverse images, having 8.3 object categories per image. Several changes were made to the AP introduced in Pascal VOC like ignoring un-annotated class, detection requirement for class and its subclass, etc. Figure~\ref{fig:openimagedataset_classes} depicts the distribution of the number of images w.r.t. to the different classes in the Open Images dataset.

\begin{figure*}[t]
\includegraphics[width=0.99\textwidth]{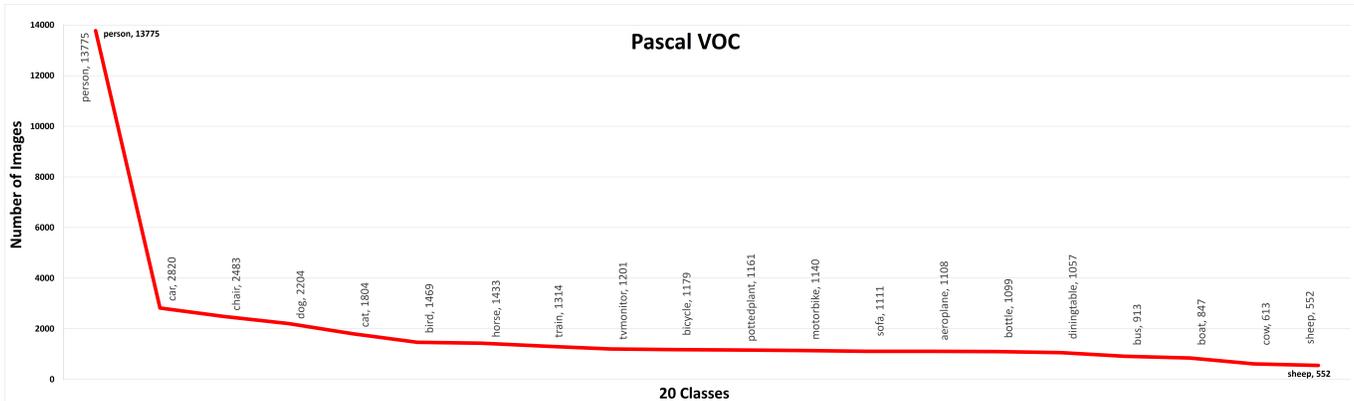}
\caption{(This image is best viewed in PDF form with magnification) Number of images for different classes annotated in the PascalVOC dataset \cite{aslam2021survey}}
\label{fig:pascalvoc_classes}
\end{figure*}

\begin{figure*}[t]
\includegraphics[width=0.99\textwidth]{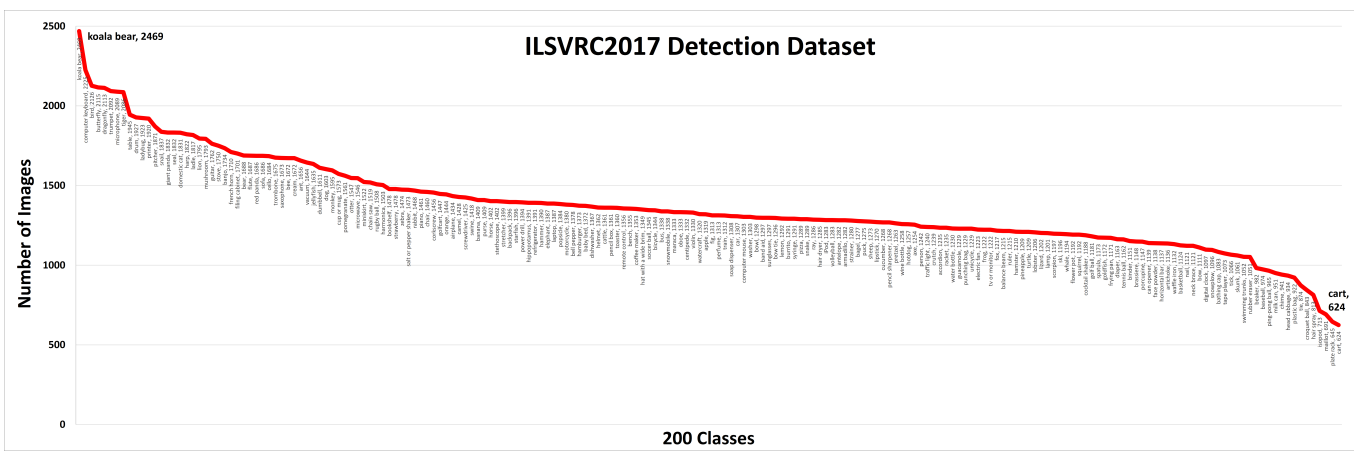}
\caption{(This image is best viewed in PDF form with magnification) Number of images for different classes annotated in the ImageNet dataset \cite{aslam2021survey}}
\label{fig:imagenet_classes}
\end{figure*}

\begin{figure*}[t]
\includegraphics[width=0.99\textwidth]{Images/OpenImageDataset-1.pdf}
\includegraphics[width=0.99\textwidth]{Images/OpenImageDataset-2.pdf}
\caption{(This image is best viewed in PDF form with magnification) Number of images for different classes annotated in the Open Images dataset \cite{aslam2021survey}}
\label{fig:openimagedataset_classes}
\end{figure*}

\subsubsection{Issues of Data Skew/Bias}

While observing Fig.~\ref{fig:pascalvoc_classes} through Fig.~\ref{fig:openimagedataset_classes}, an alert reader would certainly notice that the number of images for difference classes vary significantly in all the datasets \cite{aslam2021survey}. Three (Pascal VOC, MS-COCO, and Open Images Dataset) of the four datasets discussed above have a very significant drop in the number of images beyond the top-5 most frequent classes. As can be readily observed for Fig.~\ref{fig:pascalvoc_classes}, there are 13775 images which contain a `person' and then 2829 images which contain a `car'. The number of images for the remaining 18 classes in this dataset almost fall linearly to the  55 images of `sheep'. Similarly, for the MS-COCO dataset, the class `person' has 262465 images, and the next most-frequent class `car' has  43867 images. The downward trend continues till there are only 198 images for the class `hair drier'. A similar phenomenon is also observed in the Open Images Dataset, wherein the class `Man' is the most frequent with 378077 images, and the class `Paper Cutter' has only 3 images. This clearly represents a skew in the datasets and is bound to create a bias in the training process of any object detection model. Therefore, an object detection model trained on these skewed datasets will in all probability show better detection performance for the classes with more number of images in the training data. Although still present, this issue is slightly less pronounced in the ImageNet dataset, as can be observed from Fig.~\ref{fig:imagenet_classes} from where it can be seen that the most frequent class i.e. `koala' has  2469 images, and the least frequent class i.e. `cart' has 624 images. However, this leads to another point of concern in the ImageNet dataset: the most frequent class is for `koala' and the next most-appearing class is `computer keyboard', which are clearly not the most sought after objects in a real-world object detection scenario (where person, cars, traffic signs, etc. are of higher concern).

\subsection{Metrics}
\label{secc:metrics}

Object detectors use multiple criteria to measure the performance of the detectors viz., frames per second (FPS), precision and recall. However, mean Average Precision (\textit{mAP}) is the most common evaluation metric. Precision is derived from Intersection over Union (IoU), which is the ratio of the area of overlap and the area of union between the ground truth and the predicted bounding box. A threshold is set to determine if the detection is correct. If the IoU is more than the threshold, it is classified as True Positive while an IoU below it is classified as False Positive. If the model fails to detect an object present in the ground truth, it is termed as False Negative. Precision measures the percentage of correct predictions while the recall measure the correct predictions with respect to the ground truth. 

\begin{equation}
\begin{split}
Precision & = \frac{True\ Positive}{True\ Positive + False\ Positive}\\
 & = \frac{True\ Positive}{All\ Observations}
\end{split}
\end{equation}
\begin{equation}
\begin{split}
Recall & = \frac{True\ Positive}{True\ Positive + False\ Negative}\\
 & = \frac{True\ Positive}{All\ Ground\ Truth}
\end{split}
\end{equation}

Based on the above equation, average precision is computed separately for each class. To compare performance between the detectors, the mean of average precision of all classes, called mean average precision (\textit{mAP}) is used, which acts as a single metric for final evaluation.

\section{Backbone architectures}
\label{BAr}

Backbone architectures are one of the most important component of the object detector. These networks extract feature from the input image used by the model. Here, we have discussed some milestone backbone architectures used in modern detectors:

\subsection{AlexNet}

Krizhevsky et al. proposed AlexNet \cite{NIPS2012_c399862d}, a convolutional neural network based architecture for image classification, and won the ImageNet Large-Scale Visual Recognition Challenge (ILSVRC) 2012 challenge. It achieved a considerably higher accuracy (more than 26\%) than the contemporary models. AlexNet is composed of eight learnable layers - five convolutional and three fully connected layers. The last layer of the fully connected layer is connected to an \textit{N-way} (\textit{N}: number of classes) softmax classifier. It uses multiple convolutional kernels throughout the network to obtain features from the image. It also uses dropout and ReLU for regularization and faster training convergence respectively. The convolutional neural networks were given a new life by its reintroduction in AlexNet and it soon became the go-to technique in processing imaging data.

\subsection{VGG}

While AlexNet \cite{NIPS2012_c399862d} and its successors like \cite{zeiler_visualizing_2014} focused on smaller receptive window size to improve accuracy, Simonyan and Zisserman investigated the effects of network depth on it. They proposed VGG \cite{simonyan_very_2015}, which used small convolution filters to construct networks of varying depths. While a larger receptive field can be captured by a set of smaller convolutional filters, it drastically reduces network parameters and converges sooner. The paper demonstrated how deep network architecture (16-19 layers) can be used to perform classification and localization with superior accuracy. VGG was created by adding a stack of convolutional layers with three fully connected layers, followed by a softmax layer. The number of convolutional layers, according to the authors, can vary from 8 to 16. VGG is trained in multiple iterations; first, the smallest 11-layer architecture is trained with random initialization whose weights are then used to train larger networks to prevent gradient instability. VGG outperformed ILSVRC 2014 winner GoogLeNet \cite{szegedy_going_2014} in the single network performance category. It soon became one of the most used network backbones for object classification and detection models.

\subsection{GoogLeNet/Inception}

Even though classification networks were making inroads towards faster and more accurate networks, deploying them in real-world applications was still a long way off as they were resource-intensive. As networks are scaled for better performance, the computation cost increases exponentially. Szegedy et al. in \cite{szegedy_going_2014} postulated the wastage of computations in the network as a major reason for it. Bigger models also have a large number of parameters and tend to overfit the data. They proposed using locally sparse connected architecture instead of a fully connected one to solve these issues. GoogLeNet is thus a 22 layer deep network, made up by stacking multiple Inception modules on top of each other. Inception modules are networks that have multiple sized filters at the same level. Input feature maps pass through these filters and are concatenated and forwarded to the next layer. The network also has auxiliary classifiers in the intermediate layers to help regularize and propagate gradient. GoogLeNet showed how efficient use of computation blocks can perform at par with other parameter-heavy networks. It achieved 93.3\% top-5 accuracy on ImageNet \cite{russakovsky_imagenet_2015} dataset without external data, while being faster than other contemporary models. Updated versions of Inception like \cite{szegedy_rethinking_2016},\cite{szegedy_inception-v4_2016} were also published in the following years which further improved its performance and gave further evidence of the applications of refined sparsely connected architectures.

\subsection{ResNets}

As convolutional neural networks become deeper and deeper, Kaiming He et al. in \cite{kaiming_resnet_2016} showed how their accuracy first saturates and then degrades rapidly. They proposed the use of residual learning to the stacked layers to mitigate the performance decay. It is realized by addition of a skip connection between the layers. This connection is an element-wise addition between input and output of the block and does not add extra parameter or computational complexity to the network. A typical 34 layer ResNet \cite{kaiming_resnet_2016} is basically a large (\textit{7x7}) convolution filter followed by 16 bottleneck modules (pair of small \textit{3x3} filters with identity shortcut across them) and ultimately a fully connected layer. The bottleneck architecture can be adapted for deeper networks by stacking 3 convolutional layers (\textit{1x1,3x3,1x3}) instead of 2. Kaiming He et al. also demonstrated how the 16-layer VGG net had higher complexity than their considerably deeper 101 and 152 layer ResNet architectures while having lower accuracy. In subsequent paper, the authors proposed Resnetv2 \cite{he_identity_2016} which used batch normalization and ReLU layer in the blocks. It is more generalized and easier to train. ResNets are widely used in classification and detection backbones, and its core principles have inspired many networks (\cite{{huang_densely_2018,xie_aggregated_2017,szegedy_inception-v4_2016}}).

\subsection{ResNeXt}

The existing conventional methods of improving the accuracy of a model were by either increasing the depth or the width of the model. However, increasing any of these leads to higher model complexity and number of parameters while the gain margins diminish rapidly. Xie et al. introduced ResNeXt \cite{xie_aggregated_2017} architecture which is simpler and more efficient than other existing models. ResNeXt was inspired by the stacking of similar blocks in VGG/ResNet\cite{kaiming_resnet_2016,NIPS2012_c399862d} and “split-transform-merge” behavior of Inception module\cite{szegedy_going_2014}. It is essentially a ResNet where each ResNet block is replaced by an inception-like ResNeXt module. The complicated, tailored transformation modules from the Inception is replaced by topologically same modules in the ResNeXt blocks, making the network easier to scale and generalize. Xie et al. also emphasize that the cardinality (topological paths in the ResNeXt block) can be considered as a third dimension, along with depth and width, to improve model accuracy. ResNeXt is elegant and more concise. It achieved higher accuracy while having considerably fewer hyperparameters than a similar depth ResNet architecture. It was also the first runner up to the ILSVRC 2016 challenge. 

\begin{table}
\newcolumntype{A}{>{\centering}p{0.035\textwidth}}
\newcolumntype{B}{p{0.11\textwidth}}
\newcolumntype{C}{>{\centering\arraybackslash}p{0.06\textwidth}}
\caption{Comparison of Backbone architectures.}
\begin{tabular}{|B|c|A|C|C|C|}
 \hline
 Model & Year & Layers & Parameters\ (Million) & Top-1 acc\% & FLOPs\ (Billion)\\
 \hline
 AlexNet & 2012 & 7 & 62.4 & 63.3 & 1.5 \\
 VGG-16 & 2014 & 16 & 138.4 & 73 & 15.5 \\
 GoogLeNet & 2014 & 22 & 6.7 & - & 1.6 \\
 ResNet-50 & 2015 & 50 & 25.6 & 76 & 3.8 \\ 
 ResNeXt-50 & 2016 & 50 & 25 & 77.8 & 4.2 \\
 CSPResNeXt-50 & 2019 & 59 & 20.5 & 78.2 & 7.9 \\
 EfficientNet-B4 & 2019 & 160 & 19 & 83 & 4.2 \\
 \hline
\end{tabular}
\label{table:3}
\end{table}

\subsection{CSPNet}

Existing neural networks have shown incredible results in achieving high accuracy in computer vision tasks; however, they rely on excessive computational resources. Wang et al. believe that heavy inference computations can be reduced by cutting down the duplicate gradient information in the network. They proposed CSPNet \cite{wang_cspnet_2019} which creates different paths for the gradient flow within the network. CSPNet separates feature maps at the base layer into two parts. One part is passed through the partial convolution network block (e.g., Dense and Transition block in DenseNet \cite{huang_densely_2018} or Res(X) block in ResNeXt \cite{xie_aggregated_2017}) while the other part is combined with its outputs at a later stage. This reduces the number of parameters, increases the utilization of computation units and eases memory footprint. It is easy to implement and general enough to be applicable on other architectures like ResNet \cite{kaiming_resnet_2016}, ResNeXt \cite{xie_aggregated_2017}, DenseNet \cite{huang_densely_2018}, Scaled-YOLOv4 \cite{wang_scaled-yolov4_2020} etc. Applying CSPNet on these networks reduced computations from 10\% to 20\%, while the accuracy remained constant or improved. Memory cost and computational bottleneck is also reduced significantly with this method. It is leveraged in many state of the art detector models, while also being used for mobile and edge devices.

\begin{figure}[!]
\centering
\includegraphics[scale=.69]{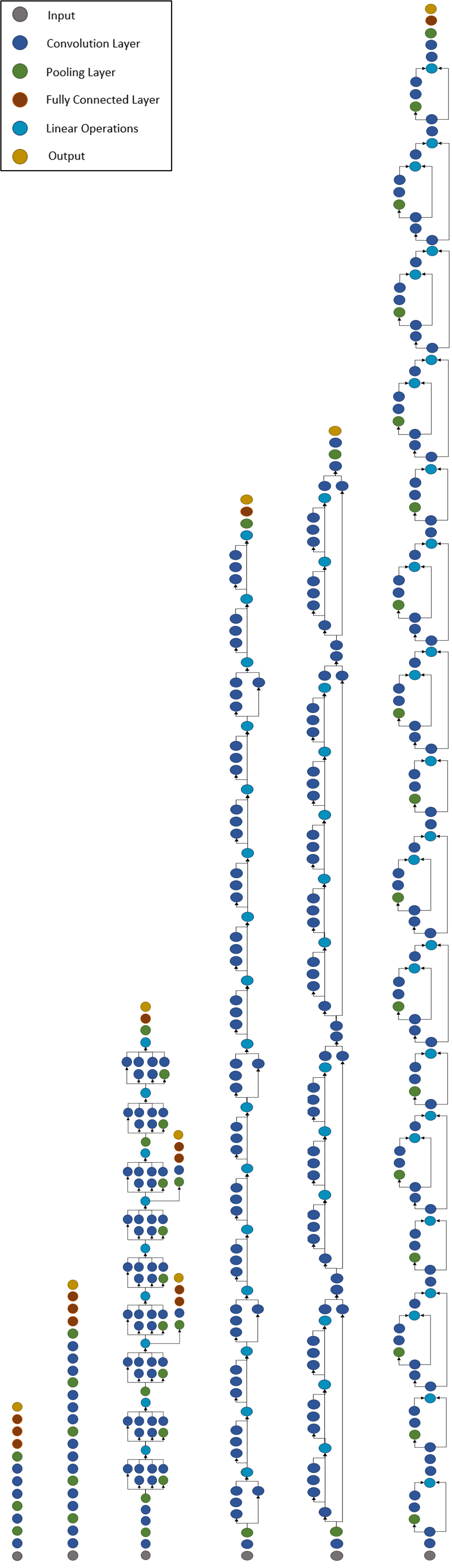}
\caption{Visualization of CNN Architectures\protect\footnotemark{}. Left to Right: AlexNet, VGG$-$16, GoogLeNet, ResNet$-$50, CSPResNeXt$-$50, EfficientNet$-$B4.}
\label{fig:NetworkArchitecture}
\end{figure}

\subsection{EfficientNet}
Tan et al. systematically studied network scaling and its effects on the model performance. They summarized how altering network parameters like depth, width and resolution influence its accuracy. Scaling any parameter individually comes with an associated cost. Increasing depth of a network can help in capturing richer and more complex features, but they are difficult to train due to vanishing gradient problem. Similarly, scaling network width will make it easier to capture fine grained features but have difficulty in obtaining high level features. Gains from increasing the image resolution, like depth and width, saturate as model scales. In the paper \cite{tan_efficientnet_2020}, Tan et al. proposed the use of a compound coefficient that can uniformly scale all three dimensions. Each model parameter has an associated constant, which is found by fixing the coefficient as $1$ and performing a grid search on a baseline network. The baseline architecture, inspired by their previous work \cite{tan_mnasnet_2018}, is developed by neural architecture search on a search target while optimizing accuracy and computations. EfficientNet is a simple and efficient architecture. It outperformed existing models in accuracy and speed while being considerably smaller. By providing a monumental increase in efficiency, it could potentially open a new era in the field of efficient networks.

\section{Object Detectors}
\label{ObD}

We have divided this review based on the two types of detectors — two-stage and single-stage detectors. However, we also discussed the pioneer work, where we briefly examine a few traditional object detectors. A network which has a separate module to generate region proposals is termed as a two-stage detector. These models try to find an arbitrary number of objects proposals in an image during the first stage and then classify and localize them in the second. As these systems have two separate steps, they generally take longer to generate proposals, have complicated architecture and lacks global context. Single-stage detectors classify and localize semantic objects in a single shot using dense sampling. They use predefined boxes/keypoints of various scale and aspect ratio to localize objects. It edges two-stage detectors in real-time performance and simpler design.
\subsection{Pioneer Work}
\subsubsection{Viola-Jones}
Primarily designed for face detection, Viola-Jones object detector \cite{viola_rapid_2001}, proposed in 2001, was an accurate and powerful detector. It combined multiple techniques like Haar-like features, integral image, Adaboost and cascading classifier. First step is to search for Haar-like features by sliding a window on the input image and uses integral image to calculate. It then uses a trained Adaboost to find the classifier of each haar feature and cascades them. Viola Jones algorithm is still used in small devices as it is very efficient and fast. 

\subsubsection{HOG Detector}
In 2005, Dalal and Triggs proposed the Histogram of Oriented Gradients (HOG) \cite{dalal_histograms_2005} feature descriptor used to extract features for object detection. It was an improvement over other detectors like \cite{{lowe_distinctive_2004,lowe_object_1999,mohan_example-based_2001,yan_ke_pca-sift_2004}}. HOG extracts gradient and its orientation of the edges to create a feature table. The image is divided into grids and the feature table is then used to create histogram for each cell in the grid. HOG features are generated for the region of interest and fed into a linear SVM classifier for detection. The detector was proposed for pedestrian detection; however, it could be trained to detect various classes.
\footnotetext{Tool Used: https://netron.app/}
\subsubsection{DPM}
Deformable Parts Model (DPM) \cite{felzenszwalb_discriminatively_2008} was introduced by Felzenszwalb et al. and was the winner Pascal VOC challenge in 2009. It used individual “part” of the object for detection and achieved higher accuracy than HOG. It follows the philosophy of \textit{divide and rule}; parts of the object are individually detected during inference time and a probable arrangement of them is marked as detection. For example, a human body can be considered as a collection of parts like head, arms, legs and torso. One model will be assigned to capture one of the parts in the whole image and the process is repeated for all such parts. A model then removes improbable configurations of the combination of these parts to produce detection. DPM based models \cite{felzenszwalb_object_2010,felzenszwalb_cascade_2010} were one of the most successful algorithms before the era of deep learning. 

\subsection{Two-Stage Detectors}
\subsubsection{\textbf{R-CNN}}
The Region-based Convolutional Neural Network (R-CNN) \cite{girshick_rich_2014} was the first paper in the R-CNN family, and demonstrated how CNNs can be used to immensely improve the detection performance. R-CNN use a class agnostic region proposal module with CNNs to convert detection into classification and localization problem. A mean-subtracted input image is first passed through the region proposal module, which produces 2000 object candidates. This module find parts of the image which has a higher probability of finding an object using Selective Search \cite{uijlings_selective_2013}. These candidates are then warped and propagated through a CNN network, which extracts a 4096-dimension feature vector for each proposal. Girshick et al. used AlexNet \cite{NIPS2012_c399862d} as the backbone architecture of the detector. The feature vectors are then passed to the trained, class-specific Support Vector Machines (SVMs) to obtain confidence scores. Non-maximum suppression (NMS) is later applied to the scored regions, based on its IoU and class. Once the class has been identified, the algorithm predicts its bounding box using a trained bounding-box regressor, which predicts four parameters i.e., center coordinates of box along with its width and height. 

R-CNN has a complicated multistage training process. The first stage is pre-training the CNN with a large classification dataset. It is then fine-tuned for detection using domain-specific images (mean-subtracted, warped proposals) by replacing of the classification layer with a randomly initialized \textit{N+1}-way classifier, \textit{N} being the number of classes, using stochastic gradient descent (SGD) \cite{lecun_backpropagation_1989}. One liner SVM and bounding box regressor is trained for each class. 

R-CNN ushered a new wave in the field of object detection, but it was slow (47 sec per image) and expensive in time and space \cite{girshick_fast_2015}. It had complex training process and took days to train on small datasets even when some of the computations were shared. 

\subsubsection{\textbf{SPP-Net}}
He et al. proposed the use of Spatial Pyramid Pooling (SPP) layer \cite{grauman_pyramid_2005} to process image of arbitrary size or aspect ratio. They realized that only the fully connected part of the CNN required a fixed input. SPP-net \cite{he_spatial_2015} merely shifted the convolution layers of CNN before the region proposal module and added a pooling layer, thereby making the network independent of size/aspect ratio and reducing the computations. The selective search \cite{uijlings_selective_2013} algorithm is used to generate candidate windows. Feature maps are obtained by passing the input image through the convolution layers of a ZF-5 \cite{zeiler_visualizing_2014} network. The candidate windows are then mapped on to the feature maps, which are subsequently converted into fixed length representations by spatial bins of a pyramidal pooling layer. This vector is passed to the fully connected layer and ultimately, to SVM classifiers to predict class and score. Similar to R-CNN \cite{girshick_rich_2014}, SPP-net has as post processing layer to improve localization by bounding box regression. It also uses the same multistage training process, except that the fine tuning is done only on the fully connected layers.

SPP-Net is considerably faster than the R-CNN model with comparable accuracy. It can process images of any shape/aspect ratio and thus, avoid object deformation due to input warping. However, as its architecture is analogous to R-CNN, it shared R-CNN’s disadvantages too like multistage training, computationally expensive and training time as well.

\subsubsection{\textbf{Fast R-CNN}}
One of the major issues with R-CNN/SPP-Net was the need to train multiple systems separately. Fast R-CNN \cite{girshick_fast_2015} solved this by creating a single end-to-end trainable system. The network takes as input an image and its object proposals. The image is passed through a set of convolution layers and the object proposals are mapped to the obtained feature maps. Girshick replaced pyramidal structure of pooling layers from SPP-net \cite{he_spatial_2015} with a single spatial bin, called RoI pooling layer. This layer is connected to 2 fully connected layer and then branches out into a \textit{N+1}-class SoftMax layer and a bounding box regressor layer, which has a fully connected layer as well. The model also changed the loss function of bounding box regressor from L2 to smooth L1 to better performance, while introducing a multi-task loss to train the network.

The authors used modified version of existing state-of-art pre-trained models like \cite{NIPS2012_c399862d}, \cite{simonyan_very_2015} and \cite{jia_caffe_2014} as backbone. The network was trained in a single step by stochastic gradient descent (SGD) and a mini-batch of 2 images. This helped the network converge faster as the back-propagation shared computations among the RoIs from the two images. 

Fast R-CNN was introduced as an improvement in speed (146x on R-CNN) while the increase in accuracy was supplementary. It simplified training procedure, removed pyramidal pooling and introduces a new loss function. The object detector, without the region proposal network, reported near real time speed with considerable accuracy.

\begin{figure*}[tb]
\includegraphics[width=1\textwidth]{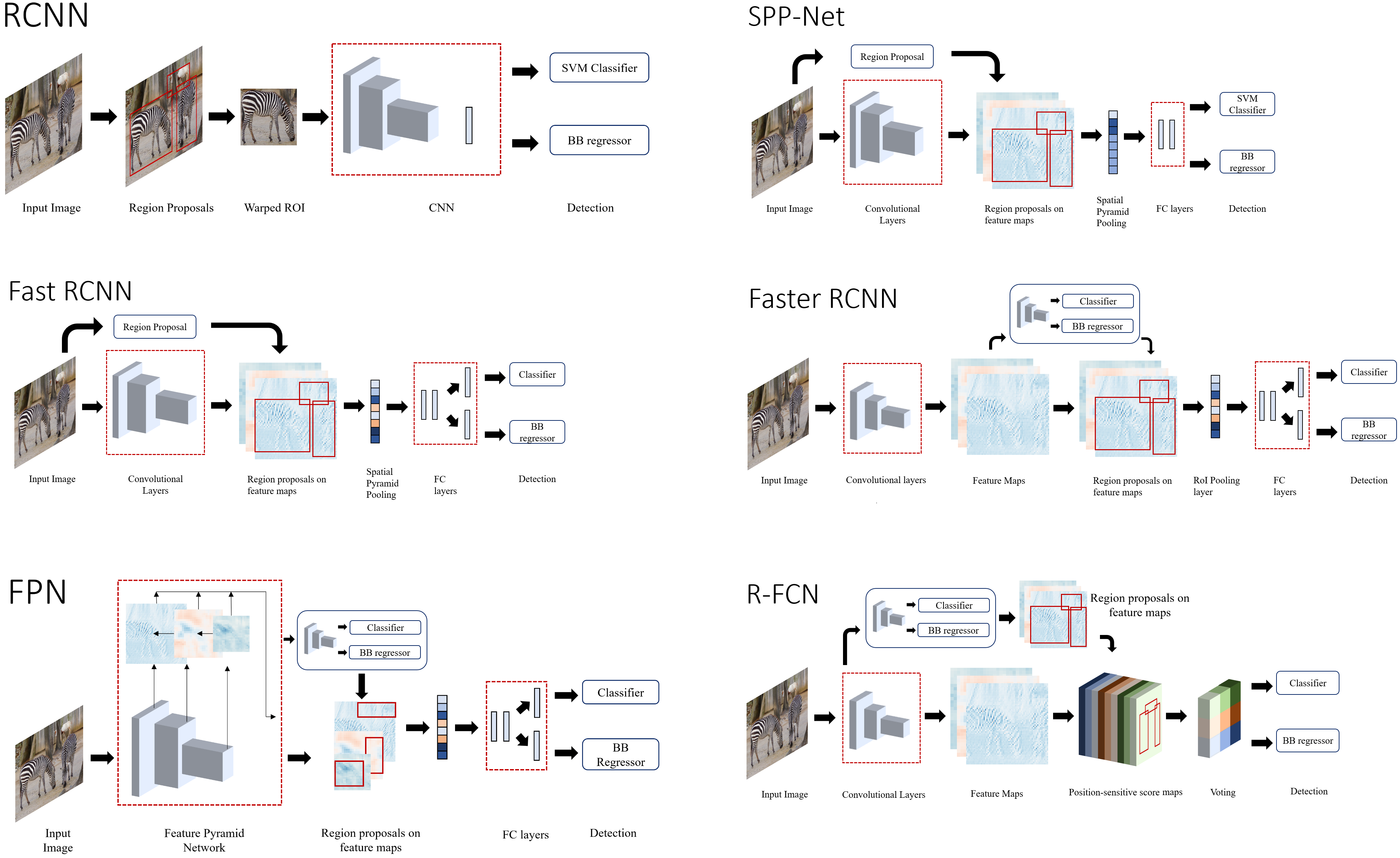}
\caption{Illustration of the internal architecture of different two stage object detectors\protect\footnotemark{\label{note2}}.}
\label{fig:twostage}
\end{figure*}

\subsubsection{\textbf{Faster R-CNN}}
Even though Fast R-CNN inched closer to real time object detection, its region proposal generation was still an order of magnitude slower (2 sec per image compared to 0.2 sec per image). Ren et al. suggested a fully convoluted network \cite{long_fully_2015} as a region proposal network (RPN) in \cite{ren_faster_2015} that takes an arbitrary input image and outputs a set of candidate windows. Each such window has an associated \textit{objectness score} which determines likelihood of an object. Unlike its predecessors like \cite{felzenszwalb_object_2010,girshick_fast_2015,kaiming_resnet_2016} which used image pyramids to solve size variance of objects, RPN introduces Anchor boxes. It used multiple bounding boxes of different aspect ratios and regressed over them to localize object. The input image is first passed through the CNN to obtain a set of feature maps. These are forwarded to the RPN, which produces bounding boxes and their classification. Selected proposals are then mapped back to the feature maps obtained from previous CNN layer in RoI pooling layer, and ultimately fed to fully connected layer, which is sent to classifier and bounding box regressor. Faster R-CNN is essentially Fast R-CNN with RPN as region proposal module. 

Training of Faster R-CNN is more convoluted, due to the presence of shared layers between two models which perform very different tasks. Firstly, RPN is pre-trained on ImageNet dataset \cite{deng_imagenet_2009} and fine-tuned on PASCAL VOC dataset \cite{everingham_pascal_2010}. A Fast R-CNN is trained from the region proposals of RPN from first step. Till this point, the networks do not have shared convolution layer. Now, we fix the convolution layers of the detector and fine-tune the unique layers in RPN. And finally, Fast R-CNN is fine-tuned from the updated RPN. 

Faster R-CNN improved the detection accuracy over the previous state-of-art \cite{girshick_fast_2015} by more than 3\% and decreased inference time by an order of magnitude. It fixed the bottleneck of slow region proposal and ran in near real time at 5 frames per second. Another advantage of having a CNN in region proposal was that it could learn to produce better proposals and thereby increase accuracy. 

\subsubsection{\textbf{FPN}}
Use of image pyramid to obtain feature pyramid (or \textit{featurized image pyramids}) at multiple levels is a common method to increase detection of small objects. Even though it increases Average Precision of the detector, the increase in the inference time is substantial. Lin et al. proposed the Feature Pyramid Network (FPN) \cite{lin2017feature}, which has a top-down architecture with lateral connections to build high-level semantic features at different scales. The FPN has two pathways, a bottom-up pathway which is a ConvNet computing feature hierarchy at several scales and a top-down pathway which upsamples coarse feature maps from higher level into high-resolution features. These pathways are connected by lateral connection by a \textit{1x1} convolution operation to enhance the semantic information in the features. FPN is used as a region proposal network (RPN) of a ResNet-101 \cite{kaiming_resnet_2016} based Faster R-CNN here. 

FPN could provide high-level semantics at all scales, which reduced the error rate in detection. It became a standard building block in future detections models and improved accuracy their accuracy across the table. It also lead to development of other improved networks like PANet \cite{liu_path_2018}, NAS-FPN \cite{ghiasi_nas-fpn_2019} and EfficientNet \cite{tan_efficientnet_2020}, which is current state of art detector. 

\subsubsection{\textbf{R-FCN}}
Dai et al. proposed Region-based Fully Convolutional Network (R-FCN) \cite{dai_r-fcn_2016} that shared almost all computations within the network, unlike previous two stage detectors which applied resource intensive techniques on each proposal. They argued against the use of fully connected layers and instead used convolutional layers. However, deeper layers in the convolutional network are translation-invariant, making them ineffective for localization tasks. The authors proposed the use of position-sensitive score maps to remedy it. These sensitive score maps encode relative spatial information of the subject and are later pooled to identify exact localization. R-FCN does it by dividing the region of interest into \textit{k x k} grid and scoring the likeliness of each cell with the detection class feature map. These scores are later averaged and used to predict the object class. R-FCN detector is a combination of four convolutional networks. The input image is first passed through the ResNet-101\cite{kaiming_resnet_2016} to get feature maps. An intermediate output (Conv4 layer) is passed to a Region Proposal Network (RPN) to identify RoI proposals while the final output is further processed through a convolutional layer and is input to classifier and regressor. The classification layer combines the generated the position-sensitive map with the RoI proposals to generate predictions while the regression network outputs the bounding box details. R-FCN is trained in a similar 4 step fashion as Faster-RCNN \cite{ren_faster_2015} whilst using a combined cross-entropy and box regression loss. It also adopts online hard example mining (OHEM) \cite{shrivastava_training_2016} during the training.

Dai et al. offered a novel method to solve the problem of translation invariance in convolutional neural networks. R-FCN combines Faster R-CNN and FCN to achieve a fast, more accurate detector. Even though it did not improve accuracy by much, but it was 2.5-20 times faster than its counterpart. 

\subsubsection{\textbf{Mask R-CNN}}
Mask R-CNN \cite{he2018mask} extends on the Faster R-CNN by adding another branch in parallel for pixel-level object instance segmentation. The branch is a fully connected network applied on RoIs to classify each pixel into segments with little overall computation cost. It uses similar basic Faster R-CNN architecture for object proposal, but adds a mask head parallel to classification and bounding box regressor head. One major difference was the use of RoIAlign layer, instead of RoIPool layer, to avoid pixel level misalignment due to spatial quantization. The authors chose the ResNeXt-101 \cite{xie_aggregated_2017} as its backbone along with the feature Pyramid Network (FPN) for better accuracy and speed. The loss function of Faster R-CNN is updated with the mask loss and as in FPN, it uses 5 anchor boxes with 3 aspect ratio. Overall training of Mask R-CNN is similar to faster R-CNN. 

Mask R-CNN performed better than the existing state of the art single-model architectures, added an extra functionality of instance segmentation with little overhead computations. It is simple to train, flexible and generalizes well in applications like keypoint detection, human pose estimation, etc. However, it was still below the real time performance ($>$30 fps).

\subsubsection{\textbf{DetectoRS}}
Many contemporary two stage detectors like \cite{ren_faster_2015,chen_hybrid_2019,cai_cascade_2017} use the mechanism of looking and thinking twice i.e. calculating object proposals first and using them to extract features to detect objects. DetectoRS \cite{qiao_detectors_2020} applies this mechanism at both macro and micro level of the network. At macro level, they propose Recursive Feature Pyramid (RFP), formed by stacking multiple feature pyramid network (FPN) with extra feedback connection from the top-down level path in FPN to the bottom-up layer. The output of the FPN is processed by the Atrous Spatial Pyramid Pooling layer (ASPP) \cite{chen_deeplab_2017} before passing it to the next FPN layer. A Fusion module is used to combine FPN outputs from different modules by creating an attention map. At micro level, Qiao et al. presented the Switchable Atrous Convolution (SAC) to regulate the dilation rate of convolution. An average pooling layer with \textit{5x5} filter and a \textit{1x1} convolution is used as a switch function to decide the rate of atrous convolution \cite{holschneider_real-time_1990}, helping the backbone detect objects at various scale on the fly. They also packed the SAC in between two global context modules \cite{hu_squeeze-and-excitation_2019} as it helps in making more stable switching. The combination of these two techniques, Recursive Feature Pyramid and Switchable Atrous Convolution results in DetectoRS. The authors incorporated the above techniques with the Hybrid Task Cascade (HTC) \cite{chen_hybrid_2019} as the baseline model and a ResNext-101 backbone. 

DetectoRS combined multiple systems to improve performance of the detector and sets the state-of-the-art for the two stage detectors. Its RFP and SAC modules are well generalized and can be used in other detection models. However, it is not suitable for real time detections as it can only process about 4 frames per second.

\subsection{Single Stage Detectors}
\subsubsection{\textbf{YOLO}}
Two stage detectors solve the object detection as a classification problem, a module presents some candidates which the network classifies as either an object or background. However, YOLO or You Only Look Once \cite{redmon_you_2016} reframed it as a regression problem, directly predicting the image pixels as objects and its bounding box attributes. In YOLO, the input image is divided into a \textit{S x S} grid and the cell where the object’s center falls is responsible for detecting it. A grid cell predicts multiple bounding boxes, and each prediction array consists of 5 elements: center of bounding box – x and y, dimensions of the box – w and h, and the confidence score. 

YOLO was inspired from the GoogLeNet model for image classification \cite{szegedy_going_2014}, which uses cascaded modules of smaller convolution networks \cite{lin_network_2014}. It is pre-trained on ImageNet data \cite{deng_imagenet_2009} till the model achieves high accuracy and then modified by adding randomly initialized convolution and fully connected layers. At training time, grid cells predict only one class as it converges better, but it is be increased during the inference time. Multitask loss, combined loss of all predicted components, is used to optimize the model. Non maximum suppression (NMS) removes class-specific multiple detections.

YOLO surpassed its contemporary single stage real time models by a huge margin in both accuracy and speed. However, it had significant shortcomings as well. Localization accuracy for small or clustered objects and limitation to number of objects per cell were its major drawbacks. These issues were fixed in later versions of YOLO \cite{redmon_yolo9000_2016,redmon_yolov3_2018,bochkovskiy_yolov4_2020}.

\subsubsection{\textbf{SSD}}
Single Shot MultiBox Detector (SSD) \cite{liu_ssd_2016} was the first single stage detector that matched accuracy of contemporary two stage detectors like Faster R-CNN \cite{ren_faster_2015}, while maintaining real time speed. SSD was built on VGG-16 \cite{simonyan_very_2015}, with additional auxiliary structures to improve performance. These auxiliary convolution layers, added to the end of the model, decrease progressively in size. SSD detects smaller objects earlier in the network when the image features are not too crude, while the deeper layers were responsible for offset of the default boxes and aspect ratios \cite{erhan_scalable_2014}.

During training, SSD match each ground truth box with the default boxes with the best jaccard overlap and train the network accordingly, similar to Multibox \cite{erhan_scalable_2014}. They also used hard negative mining and heavy data augmentation. Similar to DPM \cite{felzenszwalb_discriminatively_2008}, it utilized weighted sum of the localization and confidence loss to train the model. Final output is obtained by performing non maximum suppression. 

Even though SSD was significantly faster and more accurate than both state-of-art networks like YOLO and Faster R-CNN, it had difficulty in detecting small objects. This issue was later solved by using better backbone architectures like ResNet and other small fixes. 

\begin{figure*}[t]
\includegraphics[width=0.999\textwidth]{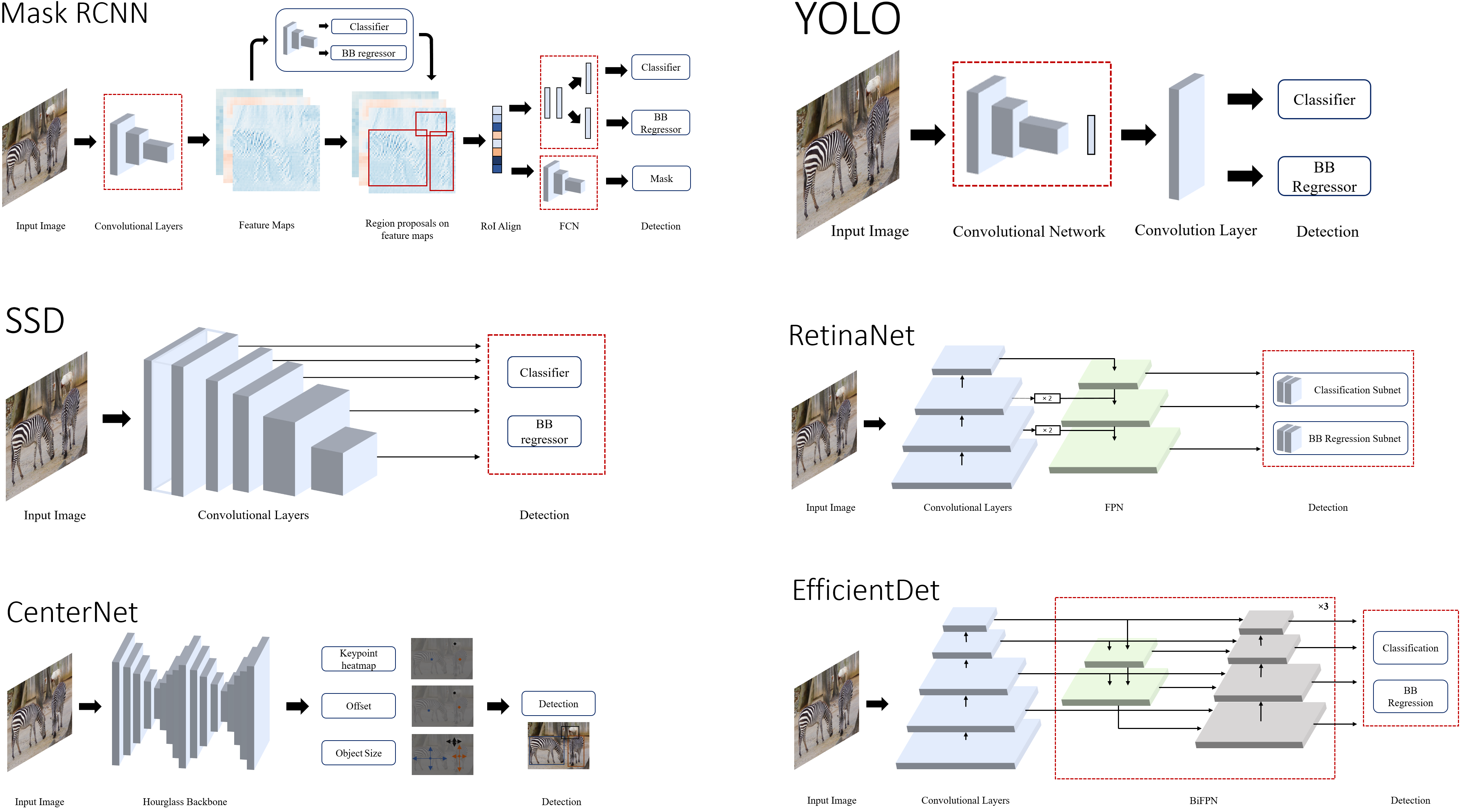}
\caption{Illustration of the internal architecture of different two and single stage object detectors\protect\footnotemark[2]{}.}
\label{fig:twostage}
\end{figure*}

\subsubsection{\textbf{YOLOv2 and YOLO9000}}
YOLOv2 \cite{redmon_yolo9000_2016}, an improvement on the YOLO \cite{redmon_you_2016}, offered an easy tradeoff between speed and accuracy while the YOLO9000 model could predict 9000 object classes in real time. They replaced the backbone architecture of GoogLeNet \cite{szegedy_going_2014} with DarkNet-19 \cite{redmon_darknet_2016}. It incorporated many impressive techniques like Batch Normalization \cite{he_delving_2015} to improve convergence, joint training of classification and detection systems to increase detection classes, removing fully connected layers to increase speed and using learnt anchor boxes to improve recall and have better priors. Redmon et al. also combined the classification and detection datasets in hierarchical structure using WordNet \cite{miller_introduction_1991}. This WordTree can be used to predict a higher conditional probability of hypernym, even when the hyponym is not classified correctly, thereby increasing the overall performance of the system. 

YOLOv2 provided better flexibility to choose the model on speed and accuracy, and the new architecture had fewer parameters. As the title of the paper suggests, it was \textit{“better, faster and stronger”} \cite{redmon_yolo9000_2016}. 

\subsubsection{\textbf{RetinaNet}}
Given the difference between the accuracies of single and two stage detectors, Lin et al. suggested that the reason single stage detectors lag is the “extreme foreground-background class imbalance” \cite{lin_focal_2017}. They proposed a reshaped cross entropy loss, called Focal loss as the means to remedy the imbalance. Focal loss parameter reduces the loss contribution from easy examples. The authors demonstrate its efficacy with the help of a simple, single stage detector, called RetinaNet \cite{lin_focal_2017}, which predicts objects by dense sampling of the input image in location, scale and aspect ratio. It uses ResNet \cite{kaiming_resnet_2016} augmented by Feature Pyramid Network (FPN) \cite{lin2017feature} as the backbone and two similar subnets - classification and bounding box regressor. Each layer from the FPN is passed to the subnets, enabling it to detect objects as various scales. The classification subnet predicts the object score for each location while the box regression subnet regresses the offset for each anchor to the ground truth. Both subnets are small FCN and share parameters across the individual networks. Unlike most previous works, the authors employ a class-agnostic bounding box regressor and found them to be equally effective.

RetinaNet is simple to train, converges faster and easy to implement. It achieved better performance in accuracy and run time than the two stage detectors. RetinaNet also pushed the envelope in advancing the ways object detectors are optimized by the introduction of a new loss function.

\subsubsection{\textbf{YOLOv3}}
YOLOv3 had “incremental improvements” from the previous YOLO versions \cite{redmon_you_2016,redmon_yolo9000_2016}. Redmon et al. replaced the feature extractor network with a larger Darknet-53 network\cite{redmon_darknet_2016}. They also incorporated various techniques like data augmentation, multi-scale training, batch normalization, among others. Softmax in classifier layer was replaced by a logistical classifier. 

Even though YOLOv3 was faster than YOLOv2 \cite{redmon_yolo9000_2016}, it lacked any ground breaking change from its predecessor. It even had lesser accuracy than an year old state-of-the-art detector \cite{lin_focal_2017}. 
\footnotetext{Features created using: https://poloclub.github.io/cnn-explainer/}

\subsubsection{\textbf{CenterNet}}
Zhou et al. in \cite{zhou_objects_2019} takes a very different approach of modelling objects as points, instead of the conventional bounding box representation. CenterNet predicts the object as a single point at the center of the bounding box. The input image is passed through the FCN that generates a heatmap, whose peaks correspond to center of detected object. It uses a ImageNet pretrained stacked Hourglass-101 \cite{newell_stacked_2016} as the feature extractor network and has 3 heads – heatmap head to determine the object center, dimension head to estimate size of object and offset head to correct offset of object point. Multitask loss of all three heads is back propagated to feature extractor while training. During inference, the output from offset head is used to determine the object point and finally a box is generated. As the predictions, not the result, are points and not bounding boxes, non-maximum suppression (NMS) is not required for post-processing.

CenterNet brings a fresh perspective and set aside years of progress in the field of object detection. It is more accurate and has lesser inference time than its predecessors. It has high precision for multiple tasks like 3D object detection, keypoint estimation, pose, instance segmentation, orientation detection and others. However, it requires different backbone architectures as general architectures that work well with other detectors give poor performance with it and vice-versa. 

\subsubsection{\textbf{EfficientDet}}
EfficientDet \cite{tan_efficientdet_2020} builds towards the idea of scalable detector with higher accuracy and efficiency. It introduces efficient multi-scale features, BiFPN and model scaling. BiFPN is bi-directional feature pyramid network with learnable weights for cross connection of input features at different scales. It improves on NAS-FPN \cite{ghiasi_nas-fpn_2019}, which required heavy training and had complex network, by removing one-input nodes and adding an extra lateral connection. This eliminates less efficient nodes and enhances high-level feature fusion. Unlike existing detectors which scale up with bigger, deeper backbone or stacking FPN layers, EfficientDet introduces a compounding coefficient which can be used to “jointly scale up all dimensions of backbone network, BiFPN network, class/box network and resolution” \cite{tan_efficientdet_2020}. EfficientDet utilizes EfficientNet \cite{tan_efficientnet_2020} as the backbone network with multiple sets of BiFPN layers stacked in series as feature extraction network. Each output from the final BiFPN layer is sent to class and box prediction network. The model is trained using SGD optimizer along with synchronized batch normalization and uses swish activation \cite{ramachandran_searching_2017}, instead of the standard ReLU activation, which is differentiable, more efficient and has better performance. 

EfficientDet achieves better efficiency and accuracy than previous detectors while being smaller and computationally cheaper. It is easy to scale, generalizes well for other tasks and is the current state-of-the-art model for single-stage object detection.

\subsubsection{\textbf{YOLOv4}}
YOLOv4 \cite{bochkovskiy_yolov4_2020} incorporated a lot of exciting ideas to design a fast and easy to train object detector that could work in existing production systems. It utilizes “bag of freebies” i.e., methods that only increase training time and do not affect the inference time. YOLOv4 utilizes data augmentation techniques, regularization methods, class label smoothing, CIoU-loss \cite{zheng_distance-iou_2019}, Cross mini-Batch Normalization (CmBN) , Self-adversarial training, Cosine annealing scheduler \cite{loshchilov_sgdr_2017} and other tricks to improve training. Methods that only affect the inference time, called “Bag of Specials”, are also added to the network, including Mish activation \cite{misra_mish_2020}, Cross-stage partial connections (CSP) \cite{wang_cspnet_2019}, SPP-Block \cite{he_spatial_2015}, PAN path aggregated block \cite{liu_path_2018} , Multi input weighted residual connections (MiWRC), etc. It also used genetic algorithm for searching hyper-parameter. It has an ImageNet pre-trained CSPNetDarknet-53 backbone, SPP and PAN block neck and YOLOv3 as detection head. 

Most existing detection algorithms require multiple GPUs to train model, but YOLOv4 can be easily trained on a single GPU. It is twice as fast as EfficientDet with comparable performance. It is the state-of-the-art for real time single stage detectors.

\subsubsection{\textbf{Swin Transformer}}
Transformers \cite{vaswani2017attention} have had a profound impact in the Natural Language Processing (NLP) domain since its inception. Its application in language models like BERT (Bidirectional Encoder Representation from Transformers) \cite{devlin2019bert}, GPT (Generative Pre-trained Transformer) \cite{radford2018improving}, T5 (Text-To-Text Transfer Transformer) \cite{2020t5} etc. have pushed the state of the art in the field. Transformers \cite{vaswani2017attention} uses the attention model to establish dependencies among the elements of the sequence and can a attend to longer context than other sequential architectures. The success of transformers in NLP sparked interest in its application in computer vision. While CNNs have been the backbone on advancement in vision, they have some inherent shortcomings like the lack of importance of global context, fixed post-training weights \cite{khan2021transformers} etc. 

Swin Transformer \cite{liu2021swin} seeks to provide a transformer based backbone for computer vision tasks. It splits the input images in multiple, non-overlapping patches and converts them into embeddings. Numerous Swin Transformer blocks are then applied to the patches in 4 stages, with each successive stage reducing the number of patches to maintain hierarchical representation. The Swin Transformer block is composed of local multi-headed self-attention (MSA) modules, based on alternating shifted patch window in successive blocks. Computation complexity becomes linear with image size in local self-attention while shifted window enables cross-window connection. \cite{liu2021swin} also shows how shifted windows increase detection accuracy with little overhead. 

Transformers present a paradigm shift from the CNN based neural networks. While its application in vision is still in a nascent stage, its potential to replace convolution from these tasks is very real. Swin Transformer achieved the state-of-the-art on MS COCO dataset, but utilises comparatively higher parameters than convolutional models.

\section{Lightweight Networks}
\label{LtW}

A new branch of research has shaped up in recent years, aimed at designing small and efficient networks for resource constrained environments as is common in Internet of Things (IoT) deployments \cite{abbas2021lightweight,karakanis2021lightweight,jadon2020low,jadon2019firenet}. This trend has percolated to the design of potent object detectors too. It is seen that although a large number of object detectors achieve excellent accuracy and perform inference in real-time, a majority of  these models require excessive computing resources and therefore cannot be deployed on edge devices. 

Many different approaches have shown exciting results in the past. Utilization of efficient components and compression techniques like pruning (\cite{lecun_optimal_1990,hassibi_advances_1993}), quantization (\cite{han_deep_2016,courbariaux_binaryconnect_2016}), hashing \cite{chen_compressing_2015}, etc. have improved the efficiency of deep learning models. Use of trained large network to train smaller models, called distillation \cite{hinton_distilling_2015}, has also shown interesting results. However in this section, we explore some prominent examples of efficient neural network design for achieving high performance on edge devices.

\subsection{SqueezeNet}
Recent advances in the field of CNNs had mostly focused on improving the state-of-the-art accuracy on the benchmark datasets, which led to an explosion of model size and their parameters. But in 2016, Iandola et al. proposed a smaller, smarter network called SqueezeNet \cite{iandola_squeezenet_2016}, which reduced the parameters while maintaining the performance. They achieved it by employing three main design strategies viz. using smaller filters, decreasing the number of input channels to \textit{3x3} filters and placing downsampling layers later in the network. The first two strategies decrease the number of parameters while attempting to preserve the accuracy and the third strategy increases the accuracy of the network. The building block of SqueezeNet is called a fire module, which consist of two layers: a squeeze layer and an expand layer, each with a ReLU activation. The squeeze layer is made up of multiple \textit{1x1} filters while the expand layer is a mix of \textit{1x1} and \textit{3x3} filters, thereby limiting the number of input channels. The SqueezeNet architecture is composed of a stack of 8 Fire modules squashed in between the convolution layers. Inspired by ResNet \cite{kaiming_resnet_2016}, SqueezeNet with residual connections was also proposed which increased the accuracy over the vanilla model. The authors also experimented with Deep Compression \cite{han_deep_2016} and achieved \textit{510$\times$} reduction in model size compared to AlexNet, while maintaining the baseline accuracy. SqueezeNet presented a good candidate for improving the hardware efficiency of the neural network architectures.

\subsection{MobileNets}
MobileNet \cite{howard_mobilenets_2017} moved away from the conventional methods of small models like shrinking, pruning, quantization or compressing, and instead used efficient network architecture. The network used depthwise separable convolution, which factorizes a standard convolution into a depthwise convolution and a \textit{1x1} pointwise convolution. A standard convolution uses kernels on all input channels and combines them in one step while the depthwise convolution uses different kernels for each input channel and uses pointwise convolution to combine inputs. This separation of filtering and combining of features reduces the computation cost and model size. MobileNet consists of 28 separate convolutional layers, each followed by batch normalization and ReLU activation function. Howard et al. also introduced the two model shrinking hyperparameters: width and resolution multiplier, in order to further improve speed and reduce size of the model. The width multiplier manipulates the width of the network uniformly by reducing the input and output channels while the resolution multiplier influences the size of the input image and its representations throughout the network. MobileNet achieves comparable accuracy to some full-fledged models while being a fraction of their size. Howard et al. also showed how it could generalize over various applications like face attribution, geolocalization and object detection. However, it was too simple and linear like VGG and therefore had fewer avenues for gradient flow. These were fixed in later iterations of this model \cite{sandler_mobilenetv2_2019,howard_searching_2019}.

\subsection{ShuffleNet}
In 2017, Zhang et al. introduced ShuffleNet \cite{zhang_shufflenet_2018}, an extremely computationally efficient neural network architecture, specifically designed for mobile devices. They recognized that many efficient networks become less effective as they scale down and purported it to be caused by expensive \textit{1x1} convolutions. In conjunction with channel shuffle, they proposed the use of group convolution to circumvent its drawback of limited information flow. ShuffleNet consists mainly of a standard convolution followed by stacks of ShuffleNet units grouped in three stages. The ShuffleNet unit is similar to the ResNet block where they use depthwise convolution in the \textit{3x3} layer and replace the \textit{1x1} layer with pointwise group convolution. The depthwise convolution layer is preceded by a channel shuffle operation. The computation cost of the ShuffleNet can be administered by two hyperparameters: group number to control the connection sparsity and scaling factor to manipulate the model size. As group numbers become large, the error rate saturates as the input channels to each group decreases and therefore may reduce the representational capabilities. ShuffleNet outperformed contemporary models (\cite{NIPS2012_c399862d,szegedy_going_2014,iandola_squeezenet_2016,howard_mobilenets_2017}) while having considerably smaller size. As the only advancement in ShuffleNet was channel shuffle, there isn’t any improvement in inference speed of the model. 

\subsection{MobileNetv2}
Improving on MobileNetv1 \cite{howard_mobilenets_2017}, Sandler et al. proposed MobileNetv2 \cite{sandler_mobilenetv2_2019} in 2018. It introduced the inverted residual with linear bottleneck, a novel layer module to reduce computation and improve accuracy. The module expands a low-dimensional representation of the input into high dimension, filters with a depthwise convolution and then projects it back to low dimension, unlike the common residual block which performs compression, convolution and then expansion operations. The MobileNetv2 contains a convolution layer followed by 19 residual bottleneck modules and subsequently two convolutional layers. The residual bottleneck module has a shortcut connection only when the stride is 1. For higher stride, the shortcut is not used because of the difference in dimensions. They also employed ReLU6 as the non-linearity function, instead of simple ReLU, to limit computations. For object detection, the authors used MobileNetv2 as the feature extractor of a computationally efficient variant of the SSD \cite{liu_ssd_2016}. This model, called SSDLite, claimed to have 8x fewer parameters than the original SSD while achieving competitive accuracy. It generalizes well over on other datasets, is easy to implement and hence, was well-received by the community. 

\subsection{PeleeNet}
Existing lightweight deep learning models like \cite{howard_mobilenets_2017,zhang_shufflenet_2018,sandler_mobilenetv2_2019} relied heavily on depthwise separable convolution, which lacked efficient implementation. Wang et al. proposed a novel efficient architecture based on conventional convolution, named PeleeNet \cite{wang_pelee_2018}, using an assortment of computation conserving techniques. PeleeNet was centered around the DenseNet \cite{huang_densely_2018} but looked at many other models for inspiration. It introduced two-way dense layers, stem block, dynamic number of channels in a bottleneck, transition layer compression and conventional post activation to reduce computation cost and increase speed. Inspired from \cite{szegedy_going_2014}, the two-way dense layer helps in getting different scales of the receptive field, making it easier to identify larger objects. To reduce information loss, a stem block was used in the same way to \cite{szegedy_inception-v4_2016,shen_dsod_2018}. They also parted way with the compression factor used in \cite{huang_densely_2018} as it hurts the feature expression and reduces accuracy. PeleeNet consists of a stem block, four stages of modified dense and transition layers, and ultimately the classification layer. The authors also proposed a real-time object detection system, called Pelee, which was based on PeleeNet and a variant of SSD \cite{liu_ssd_2016}. Its performance against the contemporary object detectors on mobile and edge devices was incremental but showed how simple design choices can make a huge difference in overall performance. 

\subsection{ShuffleNetv2}
In 2018, Ningning Ma et al. present a set of comprehensive guidelines for designing efficient network architectures in ShuffleNetv2 \cite{ma_shufflenet_2018}. They argued for the use of direct metrics like speed or latency to measure computational complexity, instead of indirect metrics like FLOPs. ShuffleNetv2 is built on four guiding principles – 1) equal width for input and output channels to minimize memory access cost, 2) carefully choosing group convolution based on the target platform and task, 3) multi-path structures achieve higher accuracy at the cost of efficiency and 4) element-wise operations like add and ReLU are computationally non-negligible. Following the above principles, they designed a new building block. It split the input into two parts by a channel split layer, followed by three convolutional layers which are then concatenated with the residual connection and passed through a channel shuffle layer. For the downsampling model, channel split is removed and residual connection has depthwise separable convolution layers. An ensemble of these blocks slotted in between a couple of convolutional layers results in ShuffleNetv2. The authors also experimented with larger models (50/162 layers) and obtained superior accuracy with considerably fewer FLOPs. ShuffleNetv2 punched above its weight and outperformed other state-of-the-art models at comparable complexity. 

\subsection{MnasNet}
With the increasing need for accurate, fast and low latency models for various edge devices, designing such a neural network is becoming more challenging than ever. In 2018, Tan et al. proposed Mnasnet \cite{tan_mnasnet_2018} designed from an automated neural architecture search (NAS) approach. They formulate the search problem as multi-object optimization aimed at both high accuracy and low latency. It also factorized the search space by partitioning the CNN into unique blocks and subsequently searching for operations and connections in those blocks separately, thereby reducing the search space. This also allowed each block to have a distinctive design, unlike the earlier models \cite{zoph_neural_2016,liu_progressive_2017,real_regularized_2019} which stacked the same blocks. The authors used RNN-based reinforcement learning agent as controller along with a trainer to measure accuracy and mobile devices for latency. Each sampled model is trained on a task to get its accuracy and run on the real devices for latency. This is used to achieve a soft reward target and the controller is updated. The process is repeated until the maximum iterations or a suitable candidate is derived. It is composed of 16 diverse blocks, some with residual connections. MnasNet was almost twice as fast as MobileNetv2 while having higher accuracy. However, like other reinforcement learning based neural architecture search models, the search time of MnasNet requires astronomical computational resources. 

\subsection{MobileNetv3}
At the heart of MobileNetv3 \cite{howard_searching_2019} is the same method used to create MnasNet \cite{tan_mnasnet_2018} with some modifications. A platform aware automated neural architecture search is performed in a factorized hierarchical search space and consequently optimized by NetAdapt \cite{yang_netadapt_2018}, which removes the underutilized components of the network in multiple iterations. Once an architecture proposal is obtained, it trims the channels, randomly initialize the weights and then fine-tunes it to improve the target metrics. The model was further modified to remove some expensive layer in the architecture and gain additional latency improvement. Howard et al. argued that the filters in the architecture are often mirrored images of each other, and that accuracy can be maintained even after dropping half of these filters. Using this technique reduced the computations. MobileNetv3 used a blend of ReLU and hard swish as activation filters, the latter is mostly employed towards the end of the model. Hard swish has no noticeable difference from the swish function but is computationally cheaper while retaining the accuracy. For different resource use cases, \cite{howard_searching_2019} introduced two models – MobileNetv3-Large and MobileNetv3-Small. MobileNetv3-Large is composed of 15 bottleneck blocks while MobileNetv3-Small has 11. It also included squeeze and excitation layer \cite{hu_squeeze-and-excitation_2019} on its building blocks. Similar to \cite{sandler_mobilenetv2_2019}, these model act as a feature detector in SSDLite and is 35\% faster than earlier iterations \cite{tan_mnasnet_2018,sandler_mobilenetv2_2019}, whilst achieving higher \textit{mAP}.

\begin{table*}
\newcolumntype{A}{>{\centering}p{0.20\textwidth}}
\newcolumntype{B}{p{0.15\textwidth}}
\newcolumntype{C}{>{\centering\arraybackslash}p{0.094\textwidth}}
\caption{Performance comparison of various object detectors on MS COCO and PASCAL VOC 2012 datasets at similar input image size.}
\begin{tabular}{|B|C|A|C|C|C|C|}
 \hline
 Model & Year & Backbone & Size & AP\textsubscript{[0.5:0.95]} & AP\textsubscript{0.5} & FPS\\ 
 \hline
 R-CNN* & 2014 & AlexNet & 224 & - & 58.50\% & $\sim$0.02\\ 
 SPP-Net* & 2015 & ZF-5 & Variable & - & 59.20\% & $\sim$0.23\\
 Fast R-CNN* & 2015 & VGG-16 & Variable & - & 65.70\% & $\sim$0.43\\
 Faster R-CNN* & 2016 & VGG-16 & 600 & - & 67.00\% & 5\\
 R-FCN & 2016 & ResNet-101 & 600 & 31.50\% & 53.20\% & $\sim$3\\
 FPN & 2017 & ResNet-101 & 800 & 36.20\% & 59.10\% & 5\\
 Mask R-CNN & 2018 & ResNeXt-101-FPN & 800 & 39.80\% & 62.30\% & 5\\
 DetectoRS & 2020 & ResNeXt-101 & 1333 & 53.30\% & 71.60\% & $\sim$4\\
 \rowcolor{lightgray!50} YOLO* & 2015 & (Modified) GoogLeNet & 448 & - & 57.90\% & 45\\
 \rowcolor{lightgray!50} SSD & 2016 & VGG-16 & 300 & 23.20\% & 41.20\% & 46\\
 \rowcolor{lightgray!50} YOLOv2 & 2016 & DarkNet-19 & 352 & 21.60\% & 44.00\% & 81\\
 RetinaNet & 2018 & ResNet-101-FPN & 400 & 31.90\% & 49.50\% & 12\\
 \rowcolor{lightgray!50} YOLOv3 & 2018 & DarkNet-53 & 320 & 28.20\% & 51.50\% & 45\\
 CenterNet & 2019 & Hourglass-104 & 512 & 42.10\% & 61.10\% & 7.8\\
 \rowcolor{lightgray!50} EfficientDet-D2 & 2020 & Efficient-B2 & 768 & 43.00\% & 62.30\% & 41.7\\
 \rowcolor{lightgray!50} YOLOv4 & 2020 & CSPDarkNet-53 & 512 & 43.00\% & 64.90\% & 31\\
 \rowcolor{lightgray!50} Swin-L & 2021 & HTC++ & - & 57.70\% & - & -\\
 \hline
 \multicolumn{7}{l}{$^{\mathrm{a}}$Models marked with * are compared on PASCAL VOC 2012, while others on MS COCO.Rows colored gray are real-time detectors ($>$30 FPS).}
\end{tabular}
\label{table:1}
\end{table*}

\begin{table}
\centering
\newcolumntype{A}{>{\centering}p{0.04\textwidth}}
\newcolumntype{B}{p{0.075\textwidth}}
\newcolumntype{C}{>{\centering\arraybackslash}p{0.06\textwidth}}
\caption{Comparison of Lightweight models.}
\begin{tabular}{|B|c|A|C|C|C|}
 \hline
 Model & Year & Top-1 Acc\% & Latency\ (ms) & Parameters\ (Million) & FLOPs\ (Million) \\
 \hline
 SqueezeNet & 2016 & 60.5 & - & 3.2 & 833 \\
 MobileNet & 2017 & 70.6 & 113 & 4.2 & 569 \\
 ShuffleNet & 2017 & 73.3 & 108 & 5.4 & 524 \\
 MobileNetv2 & 2018 & 74.7 & 143 & 6.9 & 300 \\
 PeleeNet & 2018 & 72.6 & - & 2.8 & 508 \\
 ShuffleNetv2 & 2018 & 75.4 & 178 & 7.4 & 597 \\
 MnasNet & 2018 & 76.7 & 103 & 5.2 & 403 \\
 MobileNetv3 & 2019 & 75.2 & 58 & 5.4 & 219 \\
 OFA & 2020 & 80.0 & 58 & 7.7 & 595 \\[.5ex] 
 \hline
\end{tabular}
\label{table:2}
\end{table}

\subsection{Once-For-All (OFA)}
The use of neural architecture search (NAS) for architecture design has produced state-of-the-art models in the past few years, however, they are compute expensive because of the sampled model training. Cai et al. in \cite{cai2020onceforall} proposed a novel method of decoupling model training stage and the neural architecture search stage. The model is trained only once and sub-networks can be distilled from it as per the requirements. Once-for-all (OFA) network provides flexibility for such sub-networks in four important dimension of a convolutional neural network – depth, width, kernel size and dimension. As they are nested within the OFA network and interfere with the training, progressive shrinking was introduced. First, the largest network is trained with all parameters set to maximum. Subsequently, network is fine-tuned by gradually reducing the parameter dimensions like kernel size, depth and width. For elastic kernel, a center of the large kernel is used as the small kernel. As the center is shared, a kernel transformation matrix is used to maintain performance. To vary depth, the first few layers are used and the rest are skipped from the large network. Elastic width employs a channel sorting operation to reorganize channels and uses the most important ones in smaller models. OFA achieved state-of-the-art of 80\% in ImageNet top-1 accuracy percentage and also won the 4\textsuperscript{th} Low Power Computer Vision Challenge (LPCVC) while reducing many order of magnitude of GPU training hours. It shows a new paradigm of designing lightweight models for a variety of hardware requirements.

\section{Comparative Results}
\label{CoR}

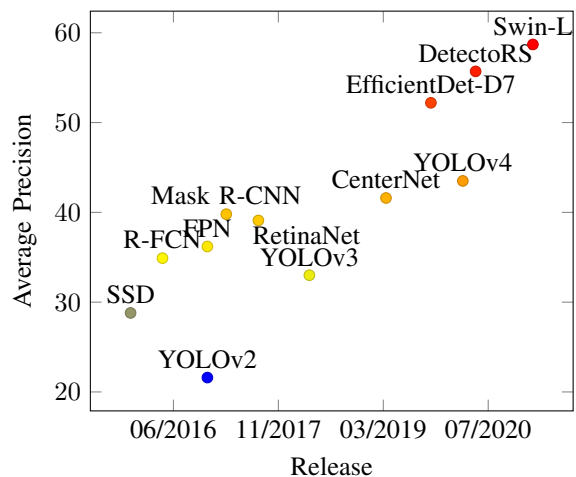
\begin{figure}
\pgfplotsset{width=8cm,compat=1.9}
\begin{tikzpicture}
  \def\MarkSize{.5em}
  \protected\def\ToWest#1{%
    \sbox0{#1}%
    \smash{%
      \rlap{%
        \kern-.2\dimexpr\wd0 + \MarkSize\relax
        \lower\dimexpr-.575em+\ht0\relax\copy0 %
      }%
    }%
    \hphantom{#1}%
  }
  \protected\def\ToEast#1{%
    \sbox0{#1}%
    \smash{%
      \rlap{%
        \kern.4\dimexpr\wd0 + \MarkSize\relax
        \lower\dimexpr.575em+\ht0\relax\copy0 %
      }%
    }%
    \hphantom{#1}%
  }

\begin{axis}[
date coordinates in=x,
xticklabel style={rotate=0,anchor=near xticklabel},
xticklabel=\month/\year,
xlabel={Release},
y tick label style={/pgf/number format/1000 sep=},
ylabel={Average Precision},
date ZERO=2009-08-18,
]
\addplot[scatter, mark=*,only marks,
        nodes near coords*={\label},visualization depends on={value \thisrow{label} \as \label},] table[x=x,y=y] {
x y label
2016-12-01 36.2 FPN
2017-03-01 39.8 Mask\ R-CNN
2015-12-01 28.8 SSD
2016-12-01 21.6 YOLOv2
2017-08-01 39.1 \ToEast{RetinaNet}
2018-04-01 33.0 YOLOv3
2019-04-01 41.6 CenterNet
2019-11-01 52.2 EfficientDet-D7
2020-04-01 43.5 YOLOv4
2020-06-01 55.7 DetectoRS
2016-05-01 34.9 R-FCN
2021-03-01 58.7 Swin-L
        }; 
\end{axis}
\end{tikzpicture}
\caption{Performance of Object Detectors on MS COCO dataset.}
\label{fig:chart}
\end{figure}

We compare the performance of both single and two stage detectors on PASCAL VOC 2012 \cite{everingham_pascal_2012} and Microsoft COCO \cite{Lin_ms_coco_2014} datasets. Performance of object detectors is influenced by a number of factors like input image size and scale, feature extractor, GPU architecture, number of proposals, training methodology, loss function etc., which makes it difficult to compare various models without a common benchmark environment. Here in table \ref{table:1}, we evaluate performance of models based on the results from their papers. Models are compared on average precision (AP) and processed frames per second (FPS) at inference time. AP\textsubscript{0.5} is the average precision of all classes when predicted bounding box has an IoU $>$ 0.5 with ground truth. COCO dataset introduced another performance metric AP\textsubscript{[0.5:0.95]}, or simply AP, which is the average AP for IoU from 0.5 to 0.95 in step size of 0.5. We intentionally compare the performances of detectors on similarly size input image, where possible, to provide a reasonable account, as authors often introduce an array of models to provide flexibility between accuracy and inference time. In fig. \ref{fig:chart}, we use only the state-of-the-art model from the possible array of object detector family of models. Lightweight models are compared in table \ref{table:2} where we compare them on ImageNet Top-1 classification accuracy, latency, number of parameters and complexity in MFLOPs. Models with MFLOPs lesser than 600 are expected to perform adequately on mobile devices.

\section{Future Trends}
\label{FuT}

Object detection has seen tremendous progress in the last decade. The algorithm have almost reached human level accuracy in some narrow domains, however it still has many exciting challenges to tackle. In this section, we discuss some of the open problem in the field of object detection. 

\textbf{AutoML}: The use of automatic neural architecture search (NAS) for determining the characteristics of object detector is already an actively growing area. We have shown some detectors designed by NAS in earlier sections, however, it is still in its nascency. Searching for an algorithm is complex and resource intensive.

\textbf{Lightweight detectors:} While lightweight networks have shown great promise by matching classification errors with the full-fledged models, they still lack in detection accuracy by more than 50\%. As more and more on-device machine learning applications are added to the market, need for small, efficient and equally accurate models will rise. 

\textbf{Weakly supervised/few shot detection:} Most of the state-of-the-art object detection models are trained on millions of bounding box annotated data, which is unscalable as annotating data requires time and resources. Ability to train on weakly supervised data, i.e. image level labelled data, cold result in considerable reduction in these costs. 

\textbf{Domain transfer:} Domain transfer refers to use of a model trained on labeled image of a particular source task on a separate, but related target task. It encourages reuse of trained model and reduces reliance on the availability of a large dataset to achieve high accuracy. 

\textbf{3D object detection:} 3D object detection is a particularly critical problem for autonomous driving. Even though models have achieved high accuracy, deployment of anything below human level performance will bring up safety concerns. 

\textbf{Object detection in video:} Object detectors are designed to perform on individual image which lack correlation between themselves. Using spatial and temporal relationship between the frames for object recognition is an open problem.

\section{Conclusion}
\label{Con}

Even though object detection has come a long way in the past decade, the best detectors are still far from saturation in performance. As its applications increase in real world, the need for lightweight models that can be deployed on mobile and embedded systems is going to increase exponentially. There has been a rising interest in this domain, but it is still an open challenge. In this paper, we have shown how two-stage and single stage detectors developed over their predecessors. While the two stage detectors are generally more accurate, they are slow and cannot be used for real-time applications like self-driving cars or security. However, this has changed in the last few year where one stage detectors are equally accurate and much faster than the former. As evident in Figure \ref{fig:chart}, Swin Transformer is the most accurate detector till date. With the current positive trend in the accuracy of detectors, we have high hopes for more accurate and faster detectors.

\balance
\bibliographystyle{./bibliography/IEEEtran}
\bibliography{./bibliography/IEEEabrv,./bibliography/Collections}

\end{document}